\documentclass[]{article}
\usepackage{amsmath, amssymb, amsfonts}
\usepackage{proceed2e}

\title{Constraint-free Graphical Model with Fast Learning Algorithm} 

\author{\bf TAKABATAKE, Kazuya\\
k.takabatake@aist.go.jp\\
Human Technology Research Institute, AIST
\And AKAHO, Shotaro\\
s.akaho@aist.go.jp\\
Human Technology Research Institute, AIST
}

\begin{document} 
 
\maketitle 
 
\begin{abstract}
In this paper, we propose a simple, versatile model for learning the structure and parameters of multivariate distributions from a data set.
Learning a Markov network from a given data set is not a simple problem, because Markov networks rigorously represent Markov properties, and this rigor imposes complex constraints on the design of the networks.
Our proposed model removes these constraints, acquiring important aspects from the information geometry.
The proposed parameter- and structure-learning algorithms are simple to execute as they are based solely on local computation at each node.
Experiments demonstrate that our algorithms work appropriately.
\end{abstract}

\section{INTRODUCTION}
The purpose of this paper is to propose a versatile mechanism for learning multivariate distributions from sets of data.
This learning mechanism is based on a simple parameter-learning algorithm and a simple structure-learning algorithm.

Markov networks are versatile tools for modeling multivariate probability distributions, because they do not require any assumptions except for the Markov property of the problem.
In this paper, we treat finite discrete systems; thus all variables take finite discrete values.

However, appropriately learning a Markov network from a given data set is not simple (Koller and Friedman, 2009).
Let us give a geometrical view of this problem.
Let $Y_i$ be the neighbors of $X_i$, and $X_{-i}$ be the variables in the network except for $X_i$.
Given a graph $G$, the Markov network that has this graph represents a manifold of distributions.
\begin{equation}
\mathcal M_G=\{p|\forall i, p(X_i|X_{-i})=p(X_i|Y_i)\}
\end{equation}
The Hammersley-Clifford theorem (Besag, 1974) proves that this manifold is identical to:
\begin{equation}
\left\{p\left|p=\frac 1Z\prod_c \phi_c(X_c)\right.\right\}
\end{equation}
where $c$ is the clique in $G$, $X_c$ denotes variables in $c$, and $Z$ denotes the normalizing constant.
\begin{figure}[h]
\begin{center}
{\scriptsize\input{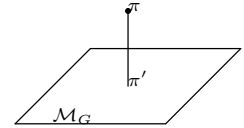}}
\end{center}
\label{fig:markovlearning}
\caption{Learning a Markov Network}
\end{figure}

Given an empirical distribution $\pi$, the role of \textit{structure-learning} is to determine a manifold $\mathcal M_G$ and the role of \textit{parameter-learning} is to determine a distribution $\pi'\in\mathcal M_G$.
If we consider maximizing likelihood (i.e., minimizing Kullback-Leibler divergence), then structure-learning algorithms should place $\mathcal M_G$ close to $\pi$, and parameter-learning algorithms should place $\pi'$ at:
\begin{equation}
\pi'=\arg\min_{p\in\mathcal M_G}KL(\pi||p).
\end{equation}
The difficulties that arise in designing learning algorithms are:
\begin{itemize}
\item $\pi'$ and $KL(\pi||\pi')$ have no closed form.
\item The problem of obtaining $\pi'$ is not decomposable; in other words, we cannot obtain $\phi_c$ independently.
\item It is intractable to find $\pi'$ in cases where the number of variables is large.
\end{itemize}
To avoid these difficulties various approaches have been attempted (see Chapter 20.9 in Koller and Friedman, 2009).
In this paper, we take a new approach that avoids these difficulties.
Firstly, we propose a new network system, named a \textit{firing process network}.
This is not a conventional graphical model, and it is obtained by relaxing the constraints of Markov networks.

In Section 2, we formulate the firing process network that the proposed learning algorithms work on, and illustrate the information geometry aspects of the firing process network.
In Section 3, we introduce the parameter-learning and structure-learning algorithms, as well as aspects of their information geometry.
We also present some information criteria that ensure the structure-learning algorithm is not overfitted.
In Section 4, we show how to draw samples from the model distribution, and also show that this model is able to draw samples from posterior distributions.
Section 5 provides experimental demonstrations that the proposed model works appropriately.

\section{FIRING PROCESS NETWORK}
\label{sec:firing process network}
\subsection{NODE}
\begin{figure}[h]
\begin{center}
{\scriptsize\input{node.tpc}}
\end{center}
\caption{Node $i$}
\end{figure}
The firing process network consists of $n$ nodes.
Herein, these nodes are indexed by the numbers $0,...,n-1$.
Each node has a variable $X_i$ and a \textit{conditional probability table} $\theta_i(X_i|Y_i)$.
Node $i$ references other nodes in the network, denoted by $Y_i$, and we call \textit{information source}.

We now assume that $Y_i=y_i$.
``Firing node $i$" means drawing a sample from distribution $\theta_i(X_i|y_i)$  (see footnote\footnote{In this paper, variables are denoted by capitals, $X,Y,Z$, and their values are denoted by lower case, $x,y,z$.
We also use shortened forms, such as $p(X|y)(=p(X|Y=y))$.}) and assigning it to the value of $X_i$.

\subsection{FORMULATION OF A FIRING PROCESS NETWORK}\label{sec:formulation}
The firing process network is formulated as follows.

$FPN=FPN(G,\theta,f_p)$: Firing process network.\\
$G$: Directed graph.\\
$X=\{X_i\}$: Nodes in $G$.\\
$Y_i$: Information source of $X_i$, i.e., nodes that have edges to $X_i$.\\
$\mathfrak y_i$: Set of node numbers included in $Y_i$.\\
$\theta=\{\theta_i\}$: Parameters.\\
$\theta_i$: Conditional probability table $\theta_i(X_i|Y_i)$.\\
$W_i$: Linear operator (=matrix) that moves a distribution $p(X)$ to the distribution $p(X_{-i})\theta(X_i|Y_i)$, i.e.
\begin{equation}
(pW_i)(X)=p(X_{-i})\theta_i(X_i|Y_i).
\label{eq:W_i}
\end{equation}
This operator represents the transition matrix caused by firing node $i$.\\
$f_p$: Either a \textit{sequential firing process} or \textit{random firing process}.
The firing processes are Markov chains.
At each time $t$, one node is chosen and fired.
Similarly to the Gibbs sampling (Gilks, Richardson and Spiegelhalter 1996), there are at least two methods to choose a node to be fired at time $t$.
One method is that we choose a node in a sequential and cyclic manner, such as $0,...,n-1,0,...,n-1,...$.
In this case, the Markov chain is a time-inhomogeneous chain because the transition matrix changes at every $t$, such as, $W_0,...,W_{n-1},W_0,...,W_{n-1},...$.
We call this Markov chain \textit{sequential firing process}.
A sequential firing processes is a time-inhomogeneous chain, however, if we observe this chain every time $n$, such as, $t=i,i+n,i+2n,...$ then the observed subsequence is a time-homogeneous chain which has the transition matrix:
\begin{equation}
W_{i\to i-1}=W_i...W_{n-1}W_0...W_{i-1}.
\label{eq:seqtransition}
\end{equation}

Another method is that we choose a node in a random manner.
Every time $t$, a random number $i$ is drawn from the distribution $c(i)$ and the node $i$ is fired\footnote{If there are no special reasons, we use uniform distribution for $c(i)$.}.
We call this Markov chain \textit{random firing process}.
A random firing process is a time-homogeneous chain, and its transition matrix is:
\begin{equation}
W=\sum_i c(i)W_i
\end{equation}
Let $D_N=\{X^0,...,X^{N-1}\}$ be a data set where $X^t$ denotes $X$ at time $t$ in a firing process, and let $p_{D_N}$ be the empirical distribution of $D_N$.
In the case of a random firing process,
\begin{equation}
\exists p^\infty,\forall p^0,\quad\lim_{t\to\infty}p^0W^t=p^\infty
\end{equation}
under the assumption of the ergodicity of $W$.
Then, by the law of large numbers,
\begin{equation}
\lim_{N\to\infty} p_{D_N}=p^\infty\quad a.s.
\label{eq:pinfty}
\end{equation}
In the case of a sequential firing process,
\begin{equation}
\forall i,\exists p^\infty_i,\forall p^0,\quad\lim_{u\to\infty}p^0W_{i\to\i-1}^u=p^\infty_i
\label{eq:pinftyi}
\end{equation}
under the assumption of the ergodicity of $W_{i\to\i-1}$.
Let $p^\infty=\frac1n p^\infty_i$, then we again get Eq.(\ref{eq:pinfty}),
because the data are considered to be drawn equally likely from $n$ time-homogeneous chains that each of their state distribution converges to $p^\infty_i$.

We define the model distribution of the firing process $\pi'$ by the limiting distribution $p^\infty$ in Eq.(\ref{eq:pinfty}), i.e.:
\begin{equation}
\pi'=p^\infty.
\end{equation}

Markov networks are a special subclass of the firing process network.
Consider the following constraints.
\begin{description}
\item[Graph constraint] All edges in $G$ are \textit{bi-directed}, i.e., if there is an edge from $X_i$ to $X_j$, then there is an edge from $X_j$ to $X_i$.
\item[Parameter constraint] There exists $\pi'\in\mathcal M_G$ such that, for all $i$, $\theta_i(X_i|Y_i)=\pi'(X_i|Y_i)$.
\end{description}
If the sequential or the random firing processes run under these constraint, they are equivalent to the Gibbs sampling and the empirical distribution of the samples converges to $\pi'$.
For a given Markov network, if we replace its edges $X_i-X_j$ with $X_i\to X_j$ and $X_j\to X_i$, then we have a firing process network that is equivalent to the given Markov network.

\subsection{INFORMATION GEOMETRY OF A FIRING PROCESS NETWORK}
\label{sec:geometry of FN}
The information geometry (Amari and Nagaoka, 1993)(Amari, 1995) illustrates important aspects of the firing process network.
We define a \textit{conditional part manifold} (see Appendix) as:
\begin{equation}
E(\theta_i)=\{p|p(X_i|X_{-i})=\theta_i(X_i|Y_i)\}.
\end{equation}
When a node $i$ is fired, the distribution of $X$ moves from $p(X)(=p(X_{-i})p(X_i|X_{-i}))$ to $p(X_{-i})\theta_i(X_i|Y_i)$; in other words, the distribution of $X$ moves from $p$ to its m-projection onto $E(\theta_i)$.

In a sequential firing process, let $\pi'_i$ be the limiting distribution (=stationary distribution) of $W_{i\to i-1}$ in Eq.(\ref{eq:seqtransition}), or in a random firing process, let $\pi'_i=\pi'W_i$.
Then, $\pi'_i$ is a distribution on $E(\theta_i)$, and $\pi$ is a mixture of them.
This implies that the model distribution of the firing process network is determined by $n$ manifolds $\{E(\theta_i)\}$, but by a single manifold such as $\mathcal M_G$ in Markov networks.
Further, note that each $\pi'_i$ has a rigorous Markov property $\pi'_i(X_i|X_{-i})=\theta_i(X_i|Y_i)$, however, these rigorous Markov properties are lost in the model distribution $\pi'$.

\section{LEARNING ALGORITHM FOR A FIRING PROCESS NETWORK}
\subsection{PARAMETER-LEARNING}\label{sec:parameter-learning}
Learning algorithms are usually designed by solving some optimization problem, i.e., to minimize or to maximize some score (e.g. likelihood).
However, we take a different approach in this paper.
Firstly, we determine a simple parameter-learning algorithm and show that this parameter-learning algorithm works appropriately under certain conditions.

Our parameter-learning algorithm is simply:
\begin{equation}
\theta_i(X_i|Y_i)=\pi(X_i|Y_i).
\end{equation}
Since $\pi(X_i|Y_i)$ is an empirical distribution of a data set, it is easily obtained by counting the samples in the data set.

In the firing process network, we compare two cases: $G$ is complete/not complete.
In the case where $G$ is complete:
\begin{equation}
\theta_i(X_i|X_{-i})=\pi(X_i|X_{-i}).
\end{equation}
Thus, the firing process (sequential,random) is equivalent to ``Gibbs sampling" and $\pi'=\pi$.
In the case where $G$ is not complete:
\begin{equation}
\theta_i(X_i|Y_i)=\pi(X_i|Y_i).
\end{equation}
We call this firing process \textit{incomplete Gibbs sampling}.
\begin{figure}[h]
\begin{center}
{\scriptsize\input{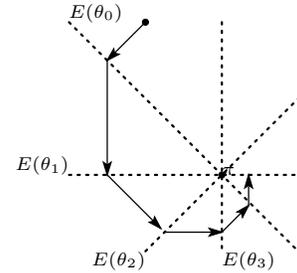}}
\end{center}
\caption{Gibbs Sampling}
\label{fig:cgs}
\end{figure}
\begin{figure}[h]
\begin{center}
{\scriptsize\input{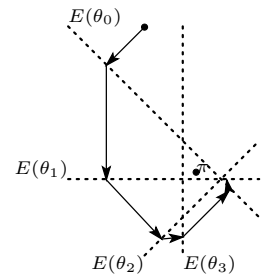}}
\end{center}
\caption{Incomplete Gibbs Sampling}
\label{fig:igs}
\end{figure}

Figs.\ref{fig:cgs} and \ref{fig:igs} illustrates the notion of information geometry of Gibbs sampling and incomplete Gibbs sampling, respectively, with the sequential firing process\footnote{In the figures in this paper, dashed lines represent e-geodesics or e-flat manifolds.}.
As described in Section \ref{sec:geometry of FN}, each time a node $i$ fires, the distribution of $X$ moves to the m-projection onto $E(\theta_i)$.
Figs.\ref{fig:cgs} and \ref{fig:igs} give us some intuition:
\begin{itemize}
\item In the Gibbs sampling, every $E(\theta_i)$ intersects at $\pi$.
Thus, the distribution of $X$ converges to $\pi$.
\item In the incomplete Gibbs sampling, each $E(\theta_i)$ does not pass $\pi$.
Thus, the distribution of $X$ does not converges to $\pi$.
However, if every $E(\theta_i)$ is close to $\pi$, then the distribution of $X$ hovers around $\pi$, thus, the model distribution $\pi'$ is close to $\pi$.
\end{itemize}
We provide more theoretical evidence for the second of these points.
For the theoretical simplicity, we treat only the random firing process in later parts of this paper.

We define a conditional part manifold $E_p^i$ and a marginal part manifold $M_p^i$:\begin{align}
&E_p^i=\{q|q(X_i|X_{-i})=p(X_i|X_{-i})\}\\
&M_p^i=\{q|q(X_{-i})=p(X_{-i})\}
\end{align}
and define the KL-divergence between a distribution $p$ and a manifold $S$:
\begin{align}
&KL(p||S)=\min_{q\in S}KL(p||q)=KL(p||pP_m(S))\\
&KL(S||p)=\min_{q\in S}KL(q||p)=KL(pP_e(S)||p).
\end{align}

\begin{figure}[h]
\begin{center}
{\scriptsize\input{fcdpq.tpc}}
\end{center}
\caption{$FCD(p||q)$}
\label{fig:fcdpq}
\end{figure}

Here, we define the following Bregman divergence (Censor and Zenios, 1997)(see Appendix) that we call \textit{full-conditional divergence}:
\begin{align}
FCD(p||q)&=B_\psi(p||q)\nonumber\\
&=\sum_i c(i)KL(p||E_q^i)\nonumber\\
&=\sum_i c(i)\left<\log\frac{p(X_i|X_{-i})}{q(X_i|X_{-i})}\right>_p,\\
\psi(p)&=\sum_i c(i)\left<\log p(X_i|X_{-i})\right>_p\nonumber\\
&=-\sum_i c(i)H_p(X_i|X_{-i})
\end{align}
where $H_p(*|*)$ denotes a conditional entropy (Cover and Thomas, 1991).
As KL-divergence is a Bregman divergence, which has a potential $-H_p(X)$,
$FCD$ is a Bregman divergence.
Thus, we can use it as a pseudo-distance.

The following inequality implies that if every conditional part manifold $E'_i$ is close to $\pi$ then the model distribution $\pi'$ is also close to $\pi$.

{\bf Upper bound of FCD)}
\begin{equation}
FCD(\pi||\pi')\le \sum_i c(i)KL(\pi||E(\theta_i)).
\label{eq:FCDbound2}
\end{equation}
Proof)
The transition matrix of the random firing process is
\begin{equation}
\sum_i c(i)P_m(E(\theta_i)).
\end{equation}
The model distribution $\pi'$ is clearly equal to the limiting distribution(=stationary distribution) of this Markov chain.
Thus:
\begin{equation}
\pi'=\pi'\sum_i c(i)P_m(E(\theta_i)).
\end{equation}
Here, we define:
\begin{equation}
\pi'_i=\pi'P_m(E(\theta_i)).
\end{equation}
Then:
\begin{equation}
\pi'=\sum_i c(i)\pi'_i.
\end{equation}
Now, Consider $KL(\pi||\pi')$.
\begin{align}
&KL(\pi||\pi')=\left<\log\frac\pi{\pi'}\right>_\pi\nonumber
=\left<\log\pi-\log\sum_i c(i)\pi'_i\right>_\pi\nonumber\\
&\quad=\left<\log\pi-\sum_i c(i)\log\pi'_i\right>_\pi\nonumber
+\sum_i c(i)\left<\log\pi'_i\right>_\pi\\
&\quad\quad-\left<\log\sum_i c(i)\pi'_i\right>_\pi\nonumber\\
&\quad=\sum_i c(i)\left<\log\frac\pi{\pi'_i}\right>_\pi-J
\label{eq:pi'pi'_ipi}
\end{align}
where
\begin{equation}
J=\left<\log\sum_i c(i)\pi'_i\right>_\pi-\sum_i c(i)\left<\log\pi'_i\right>_\pi.\end{equation}
Note that $J\ge0$, by the convexity of $-\log$.
\begin{figure}[h]
\begin{center}
{\scriptsize\input{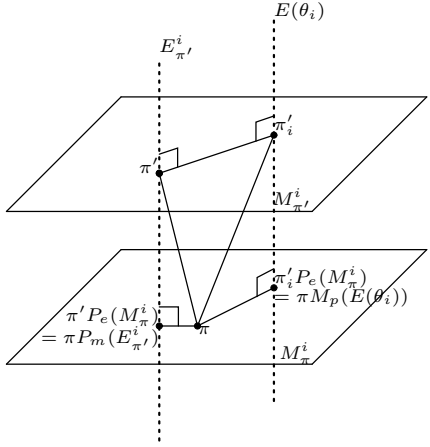}}
\end{center}
\caption{Information Geometry around $\pi',\pi'_i$ and $\pi'$}
\label{fig:pi'pi'_ipi}
\end{figure}

Fig.\ref{fig:pi'pi'_ipi} illustrates information geometric relation between $\pi',\pi'_i$ and $\pi$.
By Pythagoras' theorem in information geometry(Amari and Nagaoka, 1993)(Amari, 1995),
\begin{align}
&KL(\pi||\pi')=KL(\pi||E_{\pi'}^i)+KL(M_\pi^i||\pi')\\
&KL(\pi||\pi'_i)=KL(\pi||E(\theta_i))+KL(M_\pi^i||\pi'_i).
\end{align}
Note that:
\begin{equation}
KL(M_\pi^i||\pi')=KL(M_\pi^i||\pi'_i)
=\left<\log\frac{\pi(X_{-i})}{\pi'(X_{-i})}\right>_\pi.
\end{equation}
Subtracting:
\begin{equation}
\sum_i c(i)\left<\log\frac{\pi(X_{-i})}{\pi'(X_{-i})}\right>_\pi
\end{equation}
from Eq.(\ref{eq:pi'pi'_ipi}), we get
\begin{equation}
FCD(\pi||\pi')=\sum_i c(i)KL(\pi||E(\theta_i))-J.
\label{eq:FCDbound}
\end{equation}
Since $J\ge0$, we get the upper bound of the full-conditional divergence.
\begin{flushright}
(End of Proof
\end{flushright}

\subsection{STRUCTURE-LEARNING}
We have already determined the parameter-learning algorithm in the previous subsection, then, forming a good model depends on the structure-learning algorithm, the role of which is to determine $\{\mathfrak y_i\}$ for each node $i$.

In any machine learning algorithm, we must consider two conflicting requirements for constructing a good model:
\begin{description}
\item[A.] The model distribution $\pi'$ should be close to the data distribution $\pi$.
\item[B.] The complexity of the model should be low to avoid overfitting.
\end{description}
The previous section showed that we should place the conditional part manifold $E(\theta_i)$ close to $\pi$ for requirement A.
Since
\begin{equation}
KL(\pi||E(\theta_i))=H_\pi(X_i|Y_i)-H_\pi(X_i|X_{-i}),
\label{eq:pi2ethetai}
\end{equation}
minimizing $KL(\pi||E(\theta_i))$ is equivalent to minimizing $H_\pi(X_i|Y_i)$.
Information theory (Cover and Thomas, 1991) states that if we add a new node to $Y_i$, then $H_\pi(X_i|Y_i)$ decreases.
However, if we add a new node to $Y_i$, then the complexity of the model increases.
For an ultimate example, if we let $Y_i=X_{-i}$ then
\begin{itemize}
\item Graph $G$ becomes the complete graph.
\item The firing process becomes Gibbs sampling.
\item $\pi'=\pi$; however, the model is overfitted.
\end{itemize}

We thus use the following information criteria to determine the trade-off between the requirement A and requirement B.

\subsubsection{Node-by-node MDL/AIC}
One method of evaluating the \textit{goodness} of a model is to use some general information criteria such, as MDL (Minimum Description Length (Rissanen, 2007)) or AIC (Akaike Information Criteria (Akaike, 1974)).
However, it is difficult to apply them directly to the whole system, and therefore we apply information criteria to each node.

If we treat the conditional manifold $E(\theta_i)$ as a model manifold, then the maximum likelihood for a data set that has $N$ sample and an empirical distribution $\pi$ is:
\begin{equation}
N\times KL(\pi|E(\theta_i))=N(H_\pi(X_i|Y_i)-H_\pi(X_i|X_{-i})).
\end{equation}
The second term of the right-hand side can be neglected, because it is a constant in the situation where we select $Y_i$.
Let $k_i$ be the number of free parameters in the conditional distribution tables $\theta(X_i|Y_i)$ in the node $i$:
\begin{equation}
k_i=|Y_i|(|X_i|-1)
\end{equation}
where $|*|$ denotes the number of values that * takes.
We define:
\begin{equation}
nnMDL_i(\mathfrak y_i)=NH_\pi(X_i|Y_i)+\frac{k_i\log N}2
\label{eq:nnMDL}
\end{equation}
and call it \textit{node-by-node MDL}.

Similarly, we define:
\begin{equation}
nnAIC_i(\mathfrak y_i)=NH_\pi(X_i|Y_i)+k
\label{eq:nnAIC}
\end{equation}
and call it \textit{node-by-node AIC}.

Which information criteria to use depends on what assumptions we have about the  underlying real distribution that the data comes from.
We use $nnMDL$ in later sections.

\subsubsection{Selecting Information Source}
To find an information source $Y_i$ that minimizes $nnMDL_i(\mathfrak y_i)$, we must examine all combinations of variables in $X_{-i}$, which causes the computational costs to rise unacceptably.
Therefore, we use the following greedy algorithm (written in pseudo-Java).
{\scriptsize
\begin{quote}
\begin{tabbing}
12\=34\=\kill
$\mathfrak y_i=\emptyset$;\\
while(true)\{\\
\>$j=\arg\min_j nnMDL(\mathfrak y_i+\{j\})$;\\
\>if($nnMDL(\mathfrak y_i+\{j\})<nnMDL(\mathfrak y_i$))\{\\
\>\>$\mathfrak y_i=\mathfrak y_i+\{j\}$;\\
\>\>continue;\\
\>\}\\
\>$j=\arg\min_j nnMDL(\mathfrak y_i-\{j\})$;\\
\>if($nnMDL(\mathfrak y_i-\{j\})<nnMDL(\mathfrak y_i$))\{\\
\>\>$\mathfrak y_i=\mathfrak y_i-\{j\}$;\\
\>\>continue;\\
\>\}\\
\>break;\\
\}
\end{tabbing}
\end{quote}
}
This algorithm is similar to the forward-backward algorithms used in feature selection (Guyon and Elisseeff, 2003).

\subsection{NUMBER OF DATA AND MODEL}
In this subsection, we describe the relation between the number of data $N$ and the model distribution $\pi'$.
When the number of data is small, the second term on the right-hand side of Eq.(\ref{eq:nnMDL}) dominates $nnMDL$, and thus the number of information sources is suppressed.
\begin{figure}[h]
\begin{center}
{\scriptsize\input{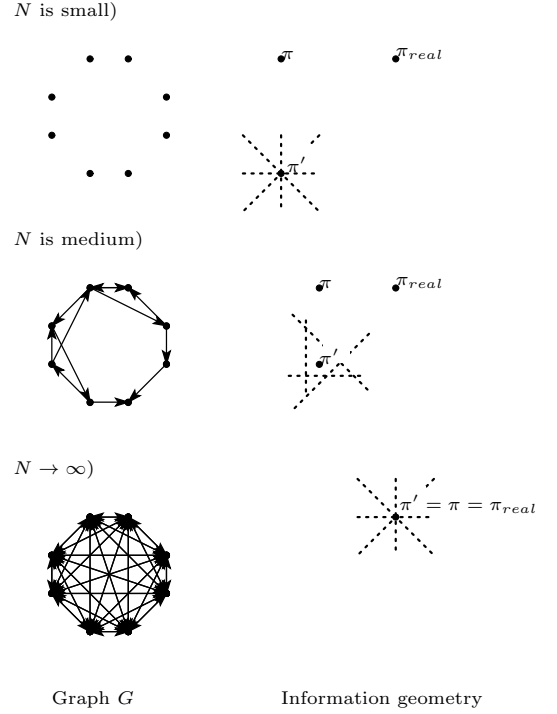}}
\end{center}
\caption{Number of Data and Proposed Model}
\label{fig:data012}
\end{figure}

Fig.\ref{fig:data012} illustrates the relation between the number of data $N$ and the proposed model.
In this figure $\pi_{real}$ denotes the underlying real distribution that the data is drawn from, and dashed lines denote $E(\theta_i)$.
Note that our goal is not to approximate $\pi$, but rather $\pi_{real}$.
In the ultimate case, $\mathfrak y_i=\emptyset$ for all nodes and $G$ has no edges, $\pi'$ is equal to the mean field approximation of $\pi$, and all $E(\theta_i)$ intersect at $\pi'$.
As $N$ increases, all $E(\theta_i)$ move toward $\pi$, and $\pi'$ approaches $\pi$.
In the case where $N\to\infty$, $\pi'=\pi=\pi_{real}$ and all $E(\theta_i)$ intersect at $\pi'=\pi=\pi_{real}$.

This behavior is reasonable because the model trusts $\pi$ when it is close to $\pi_{real}$.

\subsection{COMPUTATIONAL COST}
The key to the proposed algorithms is that computation is independently performed by each node.
This independence simplifies the situation.

Constructing a table of $\pi(X_i|Y_i)$ requires $O(N)$ computations, as it is formed by counting $N$ samples.
In addition, evaluating $H_\pi(X_i|Y_i)$ requires $O(N)$.

Adding a node to an information source of another node requires $O(nN)$ computations, as it requires the evaluation of $H_\pi(X_i|Y_i)$ at most $n$ times.
Similarly, subtracting a node from an information source of a node requires $O(nN)$.

Experiments show that subtracting nodes from information sources rarely occurs in the structure-learning.
Thus, approximately $|\mathfrak y_i|$ node additions occur during the structure-learning of node $i$.
By Eq.\ref{eq:nnMDL}, we get:
\begin{equation}
H_\pi(X_i|Y_i)+\frac{k\log N}{2N}\le nnMDL_i(\emptyset).
\end{equation}
Thus, $k=O(N/\log N)$ and:
\begin{equation}
|\mathfrak y_i|=O(\log k)=O\left(\log\frac N{\log N}\right)=O(\log N).
\end{equation}
Therefore, one node requires $O(nN\log N)$ and the whole system requires $O(n^2N\log N)$ computations for the structure-learning.

To compute $\pi'$ numerically, we must compute the eigenvector for eigenvalue 1 of the $|X|\times|X|$ transition matrix, which requires $O(|X|^3)$ computations.
Therefore, it is intractable to compute $\pi'$ for large models.

\section{SAMPLING FROM A MODEL DISTRIBUTION}
This section describes the use of our model after learning data.

This model is used as a Markov chain Monte Carlo method (Gilks, Richardson and Spiegelhalter, 1996).
We do not compute the model distribution $\pi'$ numerically, but rather draw samples from $\pi'$.
This is performed by the firing process described in Section \ref{sec:firing process network}.

\subsection{SAMPLING FROM A POSTERIOR DISTRIBUTION}
Here, we separate variables in the network into two parts: $X=(X_{-f},X_f)$.
In the Gibbs sampling, if we fix the value of $X_f$ to $x_f$ and only fire the nodes in $X_{-f}$, then we can draw samples from $\pi(X_{-f}|x_f)$.
In this paper, we call this \textit{partial sampling}
We can also conduct the partial sampling in the proposed model.

We also separate $Y_i$ into two parts: $Y=(Y_{i-f},Y_{if})$, where $Y_{i-f}$ is the variables included both in $Y_i$ and $X_{-f}$, $Y_{if}$ is the variables included both in $Y_i$ and $X_f$.

Suppose we already finished learning and obtained a network $FPN_A(G,\theta,f_p)$.
Let $FPN_B$ be the following firing process network:
\begin{itemize}
\item Nodes in $X_f$ are removed from $FPN_A$.
\item The conditional probability table of node $i$ is $\theta_i(X_i|Y_{i-f}y_{if})$ in $FPN_B$, while it is $\theta_i(X_i|Y_i)$ in $FPN_A$.
Values $y_{if}$ are fixed by $x_f$.
\end{itemize}
Then, it is clear that the partial sampling on $FPN_A$ and the normal firing process on $FPN_B$ are equivalent.
Let $\pi''(X_{-f})$ be the model distribution of $FPN_B$, and $E(\theta_{if})$ be the conditional part manifold of node $i$ in $FPN_B$, i.e.:
\begin{equation}
E(\theta_{if})=\{p(X_{-f})|p(X_i|Y_{i-f})=\theta_i(X_i|Y_{i-f}y_{if})\}
\end{equation}
Here, we can derive the following equation:
\begin{equation}
KL(\pi||E(\theta_i))=\sum_{x_f}\pi(x_f)KL(\pi(X_{-f}|x_f)||E(\theta_{if})).
\end{equation}
This equation shows that the average of $KL(\pi(X|x_f)||E(\theta_{if}))$ is equal to $KL(\pi||E(\theta_i))$.
Thus, if $KL(\pi||E(\theta_i))$ is small, then $KL(\pi(X|x_f)||E(\theta_{if}))$ is, on average, small, and the distribution of the samples drawn by the partial sampling converges to $\pi''(X_{-f})$, which is, on average, close to $\pi(X_{-f}|x_f)$.

\section{EXPERIMENTS}
\subsection{$3\times3$ ISING MODEL}
\begin{figure}[h]
{\scriptsize\input{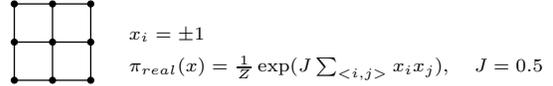}}
\caption{$3\times3$ Ising Model}
\label{fig:3x3ising}
\end{figure}
\begin{figure}[h]
\begin{center}
{\scriptsize\input{3x3result.tpc}}
\end{center}
\caption{Retrieved Structure for $3\times3$ Ising Model}
\label{fig:3x3result}
\end{figure}

We used a $3\times3$ Ising model shown in Fig.\ref{fig:3x3ising} for the first experiment.
In this case, we could compute $\pi'$ numerically, as the size of problem is small.

We used four data sets, each of which used different random seeds to draw i.i.d. (independent and identically distributed) samples from $\pi_{real}$, as shown in Fig.\ref{fig:3x3ising}.
In Fig.\ref{fig:3x3result}, the figures under the graphs are $(KL(\pi||\pi'),KL(\pi||\pi_{real}),KL(\pi'||\pi_{real}))$.
Note that in these results:
\begin{itemize}
\item $KL(\pi||\pi')\le KL(\pi||\pi_{real})$; i.e., $\pi$ is closer to $\pi'$ than to $\pi_{real}$.
\item $KL(\pi'||\pi_{real})\le KL(\pi||\pi_{real})$; i.e., $\pi_{real}$ is closer to $\pi'$ than to $\pi$.
\end{itemize}

\subsection{$5\times5$ ISING MODEL}
In learning a multivariate distribution, we often encounter the situation that $N\ll|X|$.
Therefore, we expanded the previous Ising model to $5\times5$, and similarly formed data sets by i.i.d. sampling with three different random seeds.
In this case, $|X|=2^{25}$ and it is intractable to compute $\pi'$ numerically because it would require $O(|X|^3)$ computations.
However, we can observe how the model learns the structure.

\begin{figure}[h]
{\scriptsize\input{5x5result.tpc}}
\caption{Retrieved Structure for $5\times5$ Ising Model}
\label{fig:5x5result}
\end{figure}
Fig.\ref{fig:5x5result} shows that the proposed model successfully retrieved the structure from the given data sets, and that retrieval depends on $N$ rather than $|X|$.

\subsection{ONE-DAY MOVEMENT OF STOCK PRICES}
We give the following as an example of a real-world problem.

For a one-day stock price, we define:
\begin{equation}
X_i=
\begin{cases}
1&\text{opening-price}<\text{closing-price}\\
0&\text{opening-price}\ge\text{closing-price}.
\end{cases}
\end{equation}
We followed 10 stocks and TOPIX (the overall index of stock prices on Tokyo Stock Exchange); thus, the vector $X$ consists of 11 binary variables.
We took $N=726$ samples from the real market (2009/01/05-2011/12/28) and set the proposed model to learn the distribution.

We do not know the real distribution that these samples are drawn from.
However we can observe the graph $G$ constructed by the learning algorithm.

\begin{figure}[h]
{\scriptsize\input{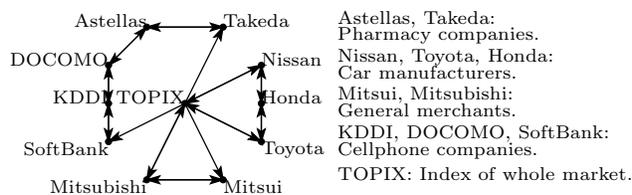}}
\caption{Learned Structure of one-day Stock Movement}
\label{fig:c11mdl}
\end{figure}
In Fig.\ref{fig:c11mdl}, every node takes other nodes in its sector or TOPIX as its information source, except for DOCOMO$\leftrightarrow$Astellas.

If we remove TOPIX from the graph, it can be noted that the nodes are separated to three groups: Domestic industry (Pharmacy, Cellphones), Exporting industry (Car manufacturers) and Importing industry (General merchants).

\section{CONCLUSION}
The important difference between conventional graphical models (Markov networks, Bayesian networks) and firing process networks is:
\begin{itemize}
\item In the conventional graphical models, the structure determines a single manifold for the entire system, and the model distribution is located on this manifold.
\item In the firing process networks, each node has a manifold respectively, thus the whole system has $n$ manifolds, and the model distribution is determined by these $n$ manifolds.
\end{itemize}
This difference makes the learning algorithms for the firing process networks simple; since each node is only responsible for its manifold, and it does not need to know what other nodes do during learning.

Future works on the proposed model will include:
\begin{itemize}
\item Comparisons with conventional learning algorithms that work on conventional graphical models.
\item Revised version of learning algorithms.
\item Expansion to continuous models.
\item Theory for sequential firing process.
\end{itemize}
\subsubsection*{References}
Akaike,~H. (1974). A new look at the statistical modeling. IEEE Trans. on Automatic Control, 19(6), pp.716-723.

Amari,~S. and Nagaoka,~H. (1993). Methods of Information Geometry.: Oxford University Press.

Amari,~S. (1995). Information geometry of the EM and em algorithms for neural networks. Neural Networks 8(9), pp.1379-1408.

Besag,~J. (1974). Spatial Interaction and Statistical Analysis of Lattice Systems. Journal of the Royal Statistical Society. Series B (Methodological) 36(2), pp.192-236

Censor,~Y. and Zenios,~S.~A. (1997). Parallel Optimization, Theory, Algorithms and Applications, Chapter 2.: Oxford University Press.

Cover,~T.~M. and Thomas,~J.~A. (1991). Elements of Information Theory, Chapter 2.: Wiley

Gilks,~W.~R., Richardson,~S. and Spiegelhalter,~D.~J. (1996). Markov Chain Monte Carlo in Practice, Chapter 3.: Chapman \& Hall.

Guyon,~I. and Elisseeff,~A. (2003). An Introduction to Variable and Feature Selection: Journal of Machine Learning Research 3, pp.1157-1182.

Koller,~D. and Friedman,~N. (2009). Probabilistic Graphical Models, Part III.: MIT Press.

Rissanen,~J. (2007). Information and Complexity in Statical Modeling: Springer.

\subsection*{Appendix}
\subsubsection*{Assumption for convergence of Markov Chain}
In this paper, we often assumed the existence of a unique limiting distribution of a Markov chain.
Here, we describe when this assumption is satisfied.
We consider only time-homogeneous Markov chains with finite state spaces here.
The following theorem for Markov chains can be found in many text books.

\textbf{Markov chain convergence theorem A})\\
There exists a unique limiting distribution $p^\infty$ for any initial distribution under assumptions:
\begin{itemize}
\item Whole system is a communicating class.
\item All states are aperiodic.
\end{itemize}
However, we cannot use this theorem directly in this paper, because the first assumption requires $\forall x, p^\infty(x)>0$.
For example, in the case that the number of data is smaller than the size of the range of $X$, the empirical distribution of the data never satisfies this assumption.
Therefore, we use the following extended version.

\textbf{Markov chain convergence theorem B})\\
There exists a unique limiting distribution $p^\infty$ for any initial distribution under assumptions:
\begin{itemize}
\item The system has a unique closed communicating class
\item All states in the communicating class are aperiodic.
\end{itemize}
It is easy to expand theorem A to theorem B, because:
\begin{itemize}
\item The system comes into the communicating class with probability 1.
\item Once the system comes into the communicating class, it will never go out.
\end{itemize}

\subsubsection*{Information Geometry of Joint Probability}\label{sec:211}
Let $X,Y$ be any stochastic variables, and $p(XY)$ be one of their joint probability.
By Bayes' rule, $p(XY)=p(X)p(Y|X)$.
Here, we call $p(X)$ \textit{marginal part} of $p$ and call $p(Y|X)$ \textit{conditional part} of $p$.
We also define two manifolds:
\begin{align*}
&M_p=\{q|q(X)=p(X)\}\\
&E_p=\{q|q(Y|X)=p(Y|X)\}.
\end{align*}
We call $M_p$ \textit{marginal part manifold} of $p$ and call $E_p$ \textit{conditional part manifold} of $p$.
These manifolds have the following properties:
\begin{itemize}
\item $M_p\cap E_p=\{p\}$
\item $M_p$ is m-flat.  $E_p$ is m-flat and e-flat.
\item $M_p\bot E_p$
\end{itemize}

Let $\mathcal M$ be any manifold, let $P_m(\mathcal M)$ be the m-projection operator onto $\mathcal M$ i.e.,
$$qP_m(\mathcal M)=\arg\min_{r\in\mathcal M} KL(q||r),$$
and let $P_e(\mathcal M)$ be the e-projection operator onto $\mathcal M$ i.e.,
$$qP_e(\mathcal M)=\arg\min_{r\in\mathcal M} KL(r||q).$$
Then:
\begin{itemize}
 \item $P_m(E_p)$ replaces the conditional part:\\
 $qP_m(E_p)=q(X)p(Y|X)$.
 \item $P_m(M_p)$ replaces the marginal part:\\
 $qP_m(M_p)=p(X)q(Y|X)$.
 \item $P_m(E_p)$ is a linear operator.
 \item $P_e(M_p)=P_m(M_p)$.
\end{itemize}

All properties in this subsection are easily proved, but we skip their proofs to save space.

\subsubsection*{Bregman Divergence}
Let $p,q$ be any vectors in $\mathbb R^N$ and $f$ be a continuously-differentiable, real-valued, and strictly convex function.
Bregman divergence is defined as:
$$B_f(p||q)=f(p)-f(q)-\nabla f(q)\cdot(p-q).$$
where $\cdot$ denotes the inner product operator.
We call $f$ \textit{potential} of $B$.
Bregman divergence has following properties:
\begin{itemize}
\item $B_f(p||q)\ge0, B_f(p||q)=0\Longleftrightarrow p=q$.
\item For any linear function $l(p)=a\cdot p+b,\\
B_{f+l}=B_f$.
\item $\min_p f(p)=f(q)=0\Longrightarrow B_f(p||q)=f(p)$
\end{itemize}

\end{document}